\documentclass{article}

\usepackage[nonatbib, final]{neurips_2022}

\usepackage[utf8]{inputenc} %
\usepackage[T1]{fontenc}    %
\usepackage{hyperref}       %
\usepackage{url}            %
\usepackage{booktabs}       %
\usepackage{amsfonts}       %
\usepackage{nicefrac}       %
\usepackage{microtype}      %
\usepackage{xcolor}         %
\usepackage{amsmath}
\usepackage[pdftex]{graphicx}

\usepackage{svg}

\title{Fully-attentive and interpretable: vision and video vision transformers for pain detection}

\author{
  Giacomo Fiorentini \\
  Utrecht University\\
  \texttt{g.fiorentini@students.uu.nl} \\
  \And
  Itir Onal Ertugrul \\
  Utrecht University\\
  \texttt{i.onalertugrul@uu.nl} \\
  \And
  Albert Ali Salah  \\
  Utrecht University\\
  \texttt{a.a.salah@uu.nl} \\
}

\begin{document}

\maketitle

\begin{abstract}
\label{abstract}
Automatic detection of facial indicators of pain has many useful applications in the healthcare domain. Vision transformers are a top-performing architecture in computer vision, with little research on their use for pain detection. In this paper, we propose the first fully-attentive automated pain detection pipeline that achieves state-of-the-art performance on binary pain detection from facial expressions. The model is trained on the UNBC-McMaster dataset, after faces are 3D-registered and rotated to the canonical frontal view. In our experiments we identify important areas of the hyperparameter space and their interaction with vision and video vision transformers, obtaining three noteworthy models. We analyze the attention maps of one of our models, finding reasonable interpretations for its predictions. We also evaluate Mixup, an augmentation technique, and Sharpness-Aware Minimization, an optimizer, with no success. Our presented models, ViT-1 (F1 score 0.55 $\pm$ 0.15), ViViT-1 (F1 score 0.55 $\pm$ 0.13), and ViViT-2 (F1 score 0.49 $\pm$ 0.04), all outperform earlier works, showing the potential of vision transformers for pain detection. Code is available at \url{https://github.com/IPDTFE/ViT-McMaster}
\end{abstract}

\section{Introduction}
\label{introduction}

The International Association for the Study of Pain defines pain as ``An unpleasant sensory and emotional experience associated with, or resembling that associated with, actual or potential tissue damage"~\cite{raja2020revised}. In Europe, one adult in five suffers from moderate to severe chronic pain, with major consequences for their lives and well-being. %
Their ability to sleep, walk, do chores, have sexual relations, live independently, and function normally feels limited or restricted~\cite{breivik2006survey}. Pain is a major healthcare problem that medical care needs to overcome.  

Pain is a ubiquitous problem for hospital care as well, with a great deal of research dedicated to pain analysis, quantification and understanding. To quantify pain, visual analogue scales (VAS)~\cite{vas}
and similar metrics are usually employed due to their convenience and simplicity. To measure pain with VAS, the patient has to point at its pain level on a horizontal scale ranging from absence to maximum pain. Unfortunately, this technique has the drawback of being subjective and easily influenced, therefore leaving much to be desired as the gold standard of pain assessment. 

Furthermore, under many circumstances patients are unable to report their pain levels, such as due to their mental and physical condition, making self-reporting techniques unreliable and widely inapplicable~\cite{ashraf2009painful,hammal2018automatic,williams2000simple}. %
In order to overcome the limitations of VAS and individual checks, automation alongside new metrics have to be employed. In this regard, facial expressions can be an important means of communication for the emotional state of a person, including their pain levels~\cite{ekman1978facial}. 

Facial expressions play an important role in communicating pain. The facial action coding system (FACS)~\cite{ekman1978facial} is a framework based on the anatomy of the facial muscles, and divides facial expression into 34 atomic components defined as action units (AU) with scores ranging from A to E depending on their intensity. While by itself this system contains no apparent information on the pain levels of the subject, the Prkachin and Solomon Pain Intensity (PSPI) score~\cite{prkachin2008structure} identifies six AUs, grouped into four actions, that contain most of the information on pain. 

These actions are brow-lowering (AU4), orbital tightening (AU6 and AU7), levator tightening (AU9 and AU10), and eye closure (AU43)~\cite{prkachin2008structure}. The PSPI score is computed by taking the highest intensity AU component of each action and summing the numerical equivalent of their intensities (ranging from 1 to 5). As AU43 (eye closure) has only one possible intensity value, PSPI is therefore a 16-point pain scale. 

\begin{align*}
PSPI = AU4 + max(AU6, AU7) + max(AU9, AU10) + AU43
\end{align*}

While the PSPI metric does not rely on self-reporting, eliminating one of the aforementioned limitations, FACS coding requires an average training time of three months, with each trained expert taking on average over two hours to code a single minute of video~\cite{clark2020facial}. In order to overcome this challenging drawback, automation is needed to predict the PSPI scores directly~\cite{kaltwang2012continuous}. Desirable properties for such automated pain detection models are spatiotemporal reasoning~\cite{ambadar2005deciphering}, robustness to occlusion and changes in the environment~\cite{rezaei2020unobtrusive,sun2019video}, explainability~\cite{tonekaboni2019clinicians}, and accuracy~\cite{walter2020automated}.
Transformer models meet many of these requirements, making them good candidates for pain assessment pipelines.
\begin{figure}[t] 
\centering
\includegraphics[width=0.8\linewidth]{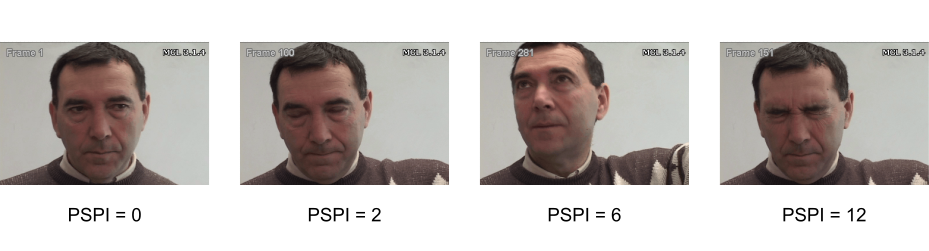}
\caption{Examples frames from the UNBC-McMaster dataset and their PSPI score labels.}
\label{fig:samples}
\end{figure}

Transformers have achieved state-of-the-art performances in multiple tasks, but only a few studies have researched their performance for pain detection~\cite{xu2021deep}. The possibility to analyze spatiotemporal relations through video transformers~\cite{arnab2021vivit}, to extract attention maps and generate interpretations intrinsic to the model~\cite{chen2021vision}~\cite{han2021transformer}, the ability to fine-tune these models on smaller datasets with good results~\cite{strudel2021segmenter}, and their state-of-the-art performance on other computer vision tasks~\cite{chen2021pre} are promising for their application towards pain assessment. 

In this study, we evaluate the performance of vision transformers (ViT) and video vision transformers (ViViT) for automated pain assessment from facial features using a fully-attentive pipeline. Training is carried out on the UNBC-McMaster dataset using PSPI labels, shown in Figure~\ref{fig:samples}, under a variety of configurations, pinpointing regions of interest in the hyperparameter space. We extract attention maps and evaluate them, finding plausible interpretations for the prediction of the model. Techniques that have been shown to boost transformer performance are evaluated and adapted to the task, attempting to maximize model performance for binary pain detection. We achieve state-of-the-art performance on the task of pain detection for the F1 score metric, demonstrating the potential of transformer models for automated pain assessment, and building foundations for future transformer research on this task. 

The contributions of this paper are:
  \begin{itemize}
  \item We propose the first fully-attentive pipeline for automated pain assessment and achieve state-of-the-art performance on binary pain detection.
  \item We identify regions of interest in the transformer hyperparameter space.
  \item We compare the performance of vision and video vision transformers.
  \item We visualize attention maps for the video vision transformer and show that pain-specific facial regions are attended.
\item We show the impact of Mixup, an augmentation technique, and SAM, an optimizer, on model performance.

\end{itemize}

\section{Related work}
\paragraph{Video Transformers} After their success in neural machine translation~\cite{vaswani2017attention}, transformers have been used as standard in several NLP tasks. Yet, their application to vision-related tasks is relatively new. Dostovitskiy et al. have proposed vision transformers (ViT) and have shown that ViT outperforms CNN once it is trained on very large databases~\cite{dosovitskiy2020image}. Recently, video vision transformers (ViViT) have been proposed to model spatiotemporal information and have been shown to achieve state-of-the-art performance on activity recognition in several settings. ViViT has outperformed earlier approaches that model spatiotemporal information~\cite{BLNET,stm} and other temporal extensions of ViT~\cite{arnab2021vivit}. 

\paragraph{Automated pain detection}

Recent work has shown that automated pain detection from facial expressions is a feasible goal. Earlier works have focused on conventional machine learning approaches such as Support Vector Machines~\cite{hammal2012automatic,werner2017automatic} and k-Nearest Neighbor (kNN)~\cite{zafar2014pain} to detect pain using a number of features extracted from face images. More recent ones have used deep learning approaches in which spatial information is learned from the face images using convolutional neural networks (CNN)~\cite{rudovic2021personalized, wang2017regularizing}. 

Modeling temporal information has been shown to be crucial as a static approach based on Relevance Vector Regression~\cite{kaltwang2012continuous} could not distinguish between eye blinks and eye closures, which are pivotal for pain intensity estimation. Recurrent convolutional neural networks (RCNN)~\cite{zhou2016recurrent} and a combination of CNNs with a long short-term memory (LSTM) networks~\cite{rodriguez2017deep} have been used to model spatiotemporal relationship among successive frames. They have shown superior performance compared to the static approaches. Inspired by these findings, we compare the performance of ViT and ViViT on automated pain detection.

\paragraph{Transformers for pain detection} Several works have shown the success of using vision transformers for facial expression recognition~\cite{ma2021facial}, and facial action unit detection~\cite{wang2021progressive}. However, their application in automated pain assessment is very scarce. To the best of our knowledge, the only existing work based on transformer technology for pain intensity estimation is by Xu and Liu~\cite{xu2021deep}. The pipeline presented in this work focuses on end-to-end pain intensity estimation and includes both a CNN and a transformer. Pain-related features are first identified and extracted from the input images by a ResNet architecture with bottleneck attention modules, then processed by a transformer model that predicts the pain intensity. The successful performance of our model on a similar task, even when only fine-tuning a pre-trained transformer, contradicts their finding that a transformer alone does not work for pain assessment.

\section{Methods}
\label{methods}

The base transformer model is pre-trained on ImageNet-21k~\cite{ridnik2021imagenet} and fine-tuned on ImageNet~\cite{ILSVRC15} at a resolution of 224x224 by HuggingFace~\cite{wolf-etal-2020-transformers} and shared under the Apache License 2.0. It employs 16x16 patches, a CLS token for classification purposes, and positional embeddings. It consists of 12 attention and 12 fully-connected layers, and employs 12 attention heads, for a total of 86 million parameters. A classification head has been added to the model, trained to interpret the CLS token and output a binary label prediction. Finally, all fully connected layers are frozen, massively reducing the training time required.

\subsection{Dataset}

The models are trained on the UNBC-McMaster dataset~\cite{lucey2011painful}, one of the most commonly used datasets for facial pain assessment. It consists of 48398 video frames from 25 patients suffering from shoulder-related pain, captured as the patients performed active and passive range-of-motion tests with each of their limbs. The dataset is extremely imbalanced, with 82.7\% of frames labeled with a PSPI score of 0 and 10.9\% having a score of 1 or 2 out of 16. For the purposes of binary classification, we divide the dataset into two categories, 0 (no-pain) and 1 (pain), the latter category including images with a PSPI score above 0. During training, the pain class is over-sampled to prevent overfitting on the majority class.

\newpage
The frames are divided into five folds, each containing samples from exactly five patients. The splits are generated with the aim of maintaining a similar number of pain samples across folds, achieved by pairing patients with the fewest and most pain samples and shuffling four patients between folds to further balance them, ensuring that each has a reasonable number of samples for the minority class. Five-fold cross-validation guarantees that the models learn to generalize painful features rather than overfitting on specific patients.

The dataset is also processed for the purposes of video transformer training - subsequent images are grouped in $2\times2$ grids and labeled according to the label of the last image of the 4-frame sequence. The use of multiple subsequent frames aims to capture the dynamics of the facial expressions, enabling the model to distinguish between the subject shutting their eyes due to pain (AU43) and blinking 
~\cite{arnab2021vivit},
and other critical dynamics of facial pain.

\subsection{3D registration}

\begin{figure}[t] \centering
   \includegraphics[width=0.8\linewidth]{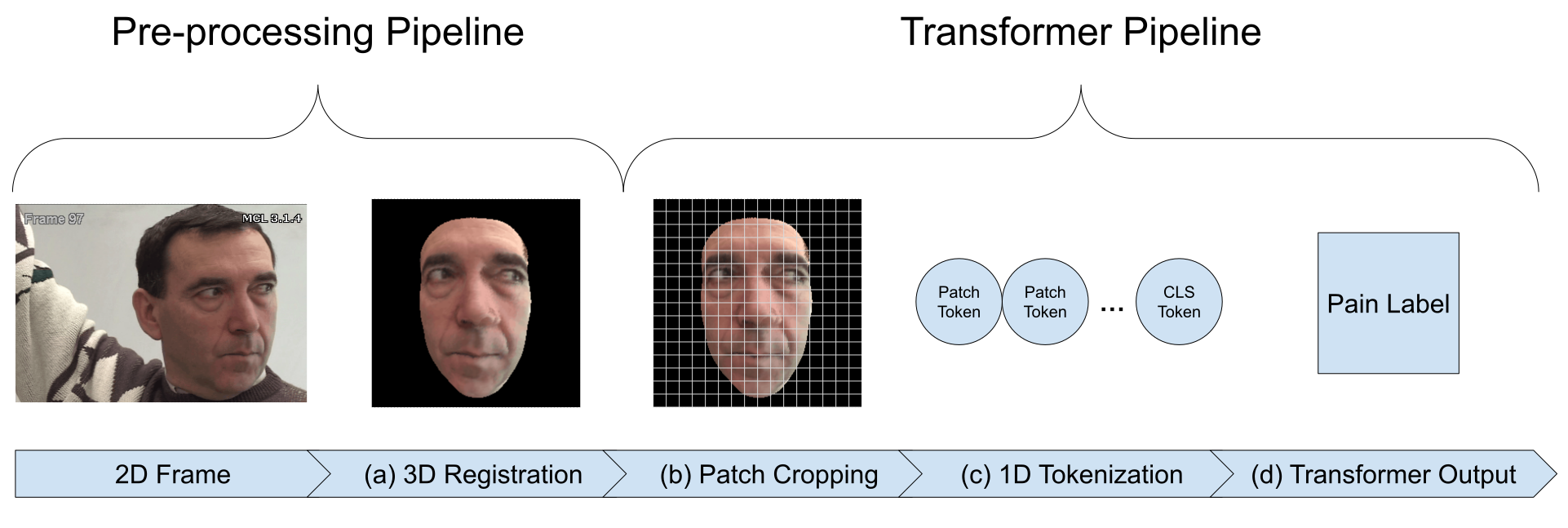}
   
    \caption{Data transformations through the pre-processing and transformer pipeline}
    \label{fig:pipeline}
\end{figure}

We perform 3D registration using PRNet~\cite{PRNET}, which gets a 2D face image as input, performs 3D registration without requiring person-specific training, and outputs a dense 3D mesh of the face. The result is achieved by regressing the UV position map, a structure that records 3D coordinates of a complete facial point cloud, from the input image. We then use Face3D tool~\cite{face3d} to rasterize 2D image from frontalized 3D facial structure generated by PRNet as shown in Figure~\ref{fig:pipeline}a.

After this step, semantic correspondence is established across frames and subjects. Consequently, visual words used in vision transformers are aligned as given in Figure \ref{fig:pipeline}b.

\subsection{Experiments}
\label{experiments}

To determine the performance of vision and video transformers in automated pain assessment, we conduct two sets of experiments.
The first set of experiments, using vision transformers, consists in tuning a single hyperparameter and saving the best performing value to be used while tuning the next parameter. The second set of experiments, identically structured, is carried out using video transformers on $2\times2$ image grids.

Due to the extreme imbalance of the labels, we evaluate the performance of the model using the F1 score on the minority class. This way, we ensure that the model prioritizes performance on the more difficult task of pain detection. Furthermore, earlier studies carried out on this task used the F1 score metric, making it possible to compare results.

We have tested 14 possible configurations, the first six seek the optimal number of unfrozen attention layers for the transformer model, then four to determine the optimal learning rate of the Adam optimizer, one to quantify the effects of the Sharpness-Aware Minimization in combination with Adam, and three for the impact of the Mixup augmentation~\cite{mixup} on the performance of the transformer. All 14 configurations have been tested separately for the single-image and the $2\times2$ grid datasets, with the rationale that the use of vision or video transformers is unlikely to be independent of each individual hyperparameter.
\newpage
First, all the fully-connected layers are kept frozen, leaving 12 attention layers to be fine-tuned. However, while too few layers cannot be effectively fine-tuned on a specific task, a higher number does not necessarily lead to a better performance~\cite{elsa}, necessitating the model to be evaluated with varying amounts of unfrozen layers. Next, the learning rate of the Adam optimizer is tuned, ahead of the introduction of the Sharpness-Aware Minimization (SAM) optimizer.

Transformer models work best with large amounts of data, nevertheless, this weakness might be mitigated with techniques such as SAM~\cite{sam} and Mixup~\cite{mixup}. The SAM optimizer works in conjunction with the original optimizer, in our case Adam, to prevent the model from converging to sharp local minima. While it could potentially reduce overfitting on the small UNBC-McMaster dataset, it also requires a second forward-backward pass, almost doubling the training time required. 

Mixup takes a percentage of the dataset and generates hybrid images by blending frames with distinct labels. For example,  a painful frame labeled [1,0] is combined with a painless frame labeled [0,1], with a blending value of 0.2. The resulting frame is labeled [0.8, 0.2], and consists of the sum of the pixel intensity values from the painful and painless picture, the former at 80\% opacity, and the latter at 20\%. 

To better integrate Mixup with the pre-processed UNBC-McMaster dataset, one further restriction is applied, allowing only images from the same patient to be combined for the experiment. The intensity of Mixup can be adjusted through its $\alpha$ parameter, causing images to be increasingly hybrid, and has been configured according to previous research on Mixup and transformers~\cite{mixupandtransformers}. Although these samples could allow for a more nuanced and linear function of pain for the model to learn from, they might also result too noisy and unnatural compared to other samples, further degrading the already limited data available.

\section{Results}
\label{Results}

Preliminary experiments have shown reasonable values for various parameters such as learning rate (2E-04), batch size (16), and number of epochs (1). 
Other important parameters for the initial training of the model are the drop-out rate before the classification head (0.10), $\beta$ values (0.9, 0.999) and $\epsilon$ (1e-08) of the Adam optimizer, weight decay (0), and the $\rho$ (0.05) of the SAM optimizer.

\subsection{Number of unfrozen attention layers}

\begin{figure}[h] \centering
\includegraphics[width=0.8\linewidth]{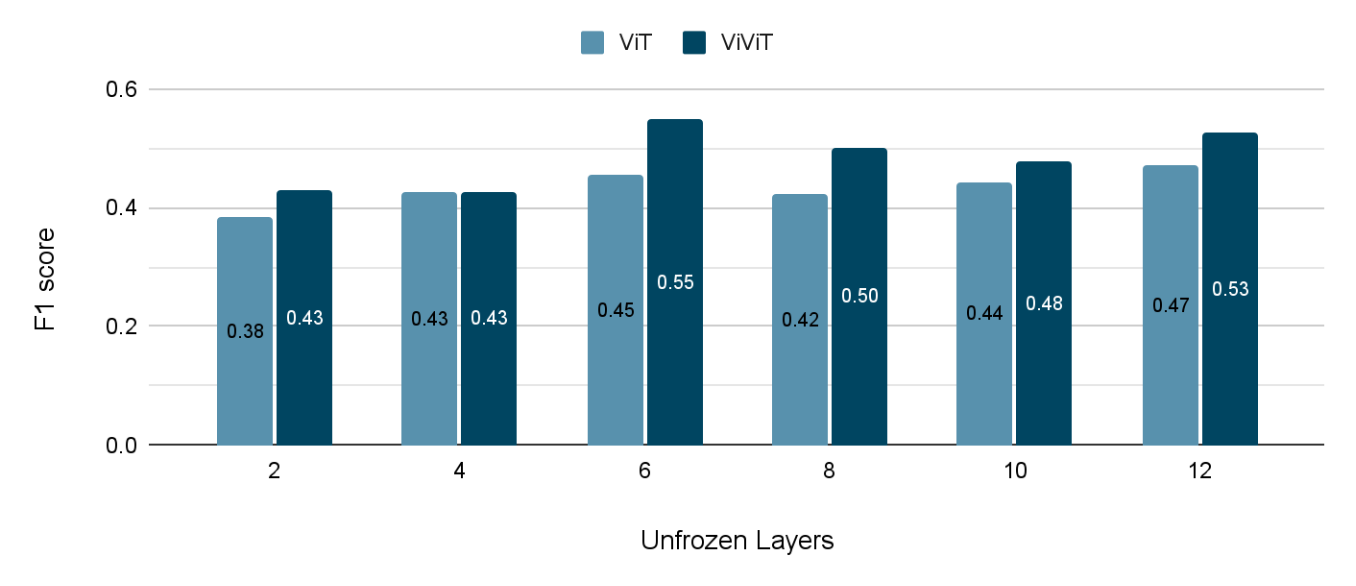}

\caption{Performance (F1 score) of ViT and ViViT models with different numbers of unfrozen (fine-tuned) attention layers.}
\label{fig:layers}
\end{figure}

The first step of the experimentation consists in identifying the optimal number of unfrozen layers. The results can be seen in Figure~\ref{fig:layers}. In total, 12 models are trained for the vision and video vision transformer with multiples of two as the number of layers, from 2 to 12. For ViT, fine-tuning 12 layers performs best (F1 score 0.47) while fine-tuning 6 layers achieves the second best performance (F1 score 0.45). For ViViT, fine-tuning 6 layers performs best (F1 score 0.55), while fine-tuning 12 layers achieves the second best performance (F1 score 0.53).
\subsection{Learning Rate}

\begin{figure}[h] \centering 
  \includegraphics[width=0.8\linewidth]{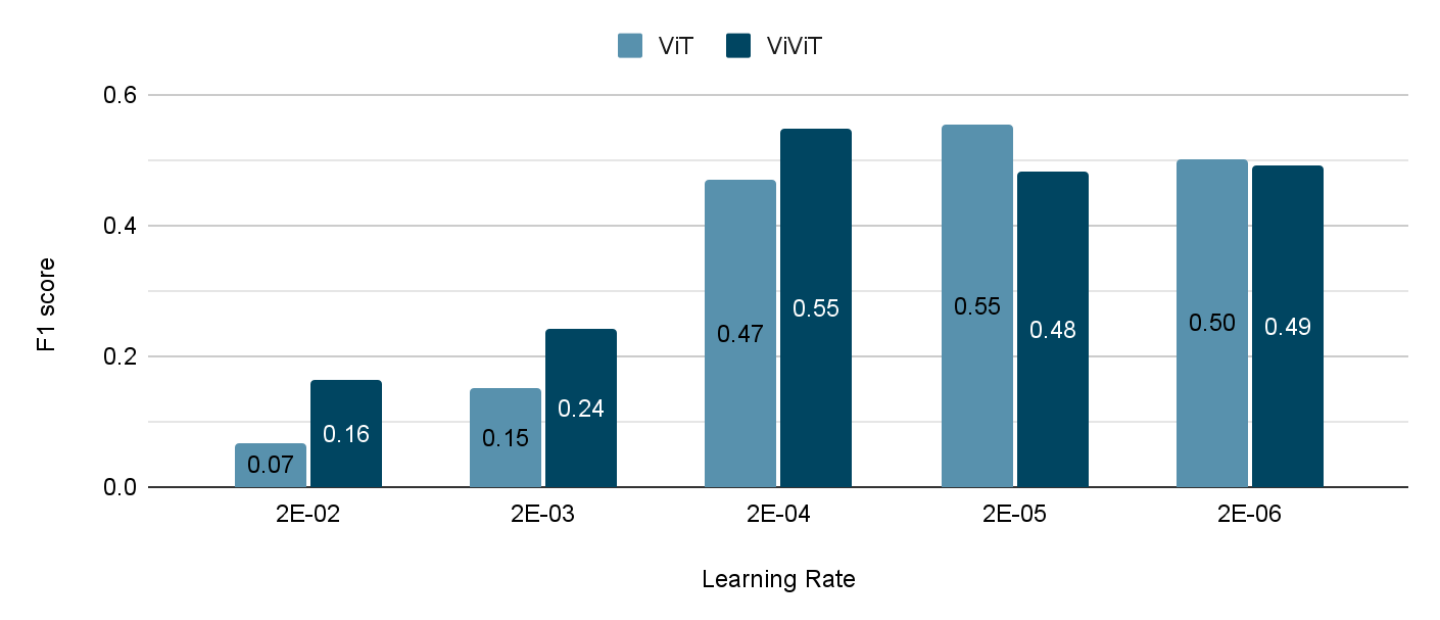}
  
\caption{Performance (F1 score) of ViT and ViViT models with different learning rates.}
\label{fig:learningrate}
\end{figure}

For the second step, a large range of learning rates is tested to identify regions of interest in the hyperparameter space. The initial learning rate of 0.0002 is both increased and decreased tenfold and a hundredfold. Performances of the resulting models can be seen in Figure~\ref{fig:learningrate}. 

\begin{figure}[h] \centering
    \includegraphics[width=0.8\linewidth]{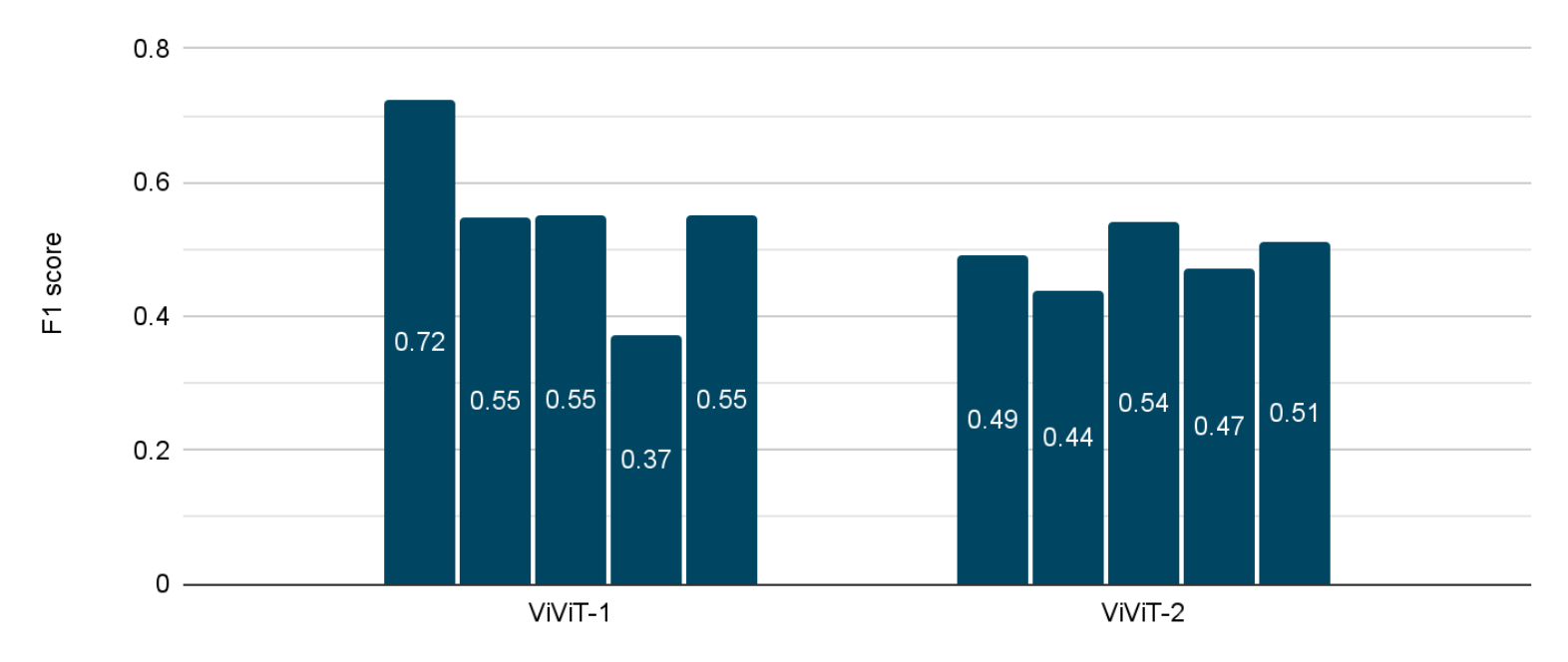}
    
\caption{Performance (F1 score) per fold of the two best performing ViViT models.}
\label{fig:learningratestd}
\end{figure}

ViT performance peaks with a learning rate of 2E-05 (F1 score 0.55, model ViT-1), followed by 2E-06 (F1 score 0.50). ViViT performs the best with a learning rate of 2E-04 (F1 score 0.55, model ViViT-1), and obtains its second best performance with a learning rate of 2E-06 (F1 score 0.49, model ViViT-2). However, a peculiar trait emerges from the latter model, an extremely low standard deviation across folds of the F1 score as visible in Table~\ref{tab:results}. The models ViT-1 (al = 12, lr = 2E-05) and ViViT-1 (al = 6, lr = 2E-04) are the best performing models of their type across all experiments, while ViViT-2 (al = 6, lr = 2E-06) is the second best performing video vision transformer and has a uniquely low standard deviation. Visible in Figure~\ref{fig:learningratestd} is a comparison of the best two ViViT models, showcasing the good performance across all folds for ViViT-2 compared to ViViT-1.
  
\subsection{Sharpness-Aware Minimization}

\begin{figure}[h] \centering
    \includegraphics[width=0.6\linewidth]{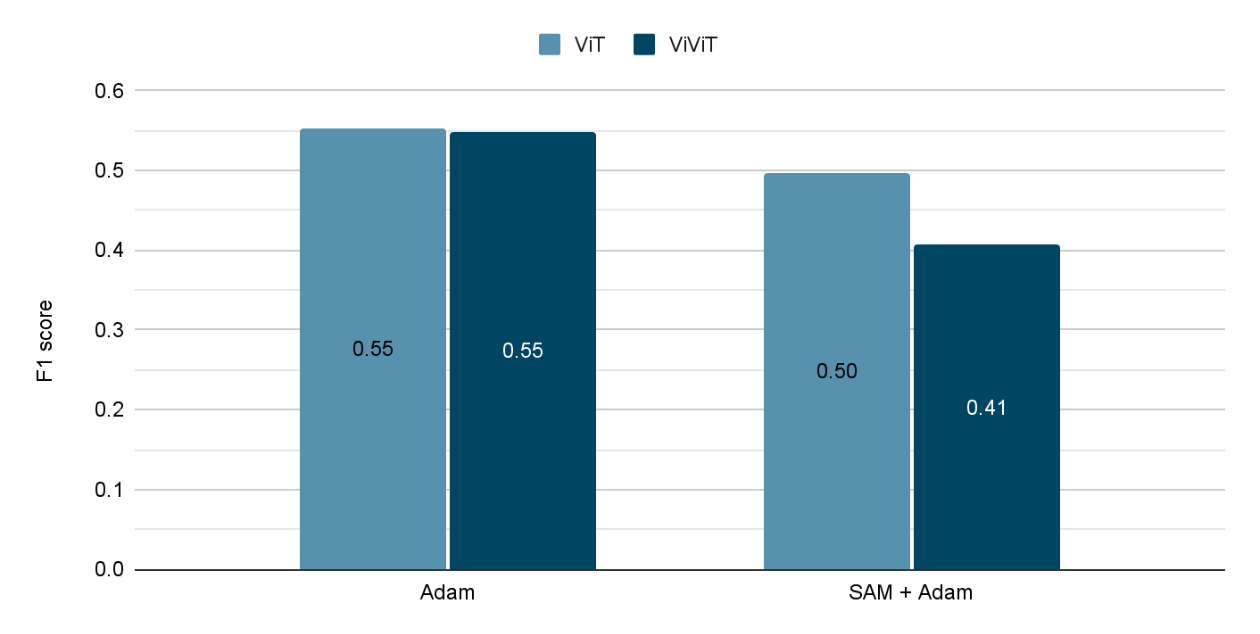}
    
 \caption{Performance (F1 score) of ViT and ViViT models optimized with and without the Sharpness-Aware Minimization (SAM).}
\label{fig:SAM} 
\end{figure}

The third step of experimentation introduces SAM to the model's training, however, this addition not only almost doubles the training time necessary but also worsens the performance of ViViT (F1 score  0.40) and ViT (F1 score 0.50), as shown in Figure~\ref{fig:SAM}. 

\subsection{Mixup}

\begin{figure}[h] \centering
\includegraphics[width=0.8\linewidth]{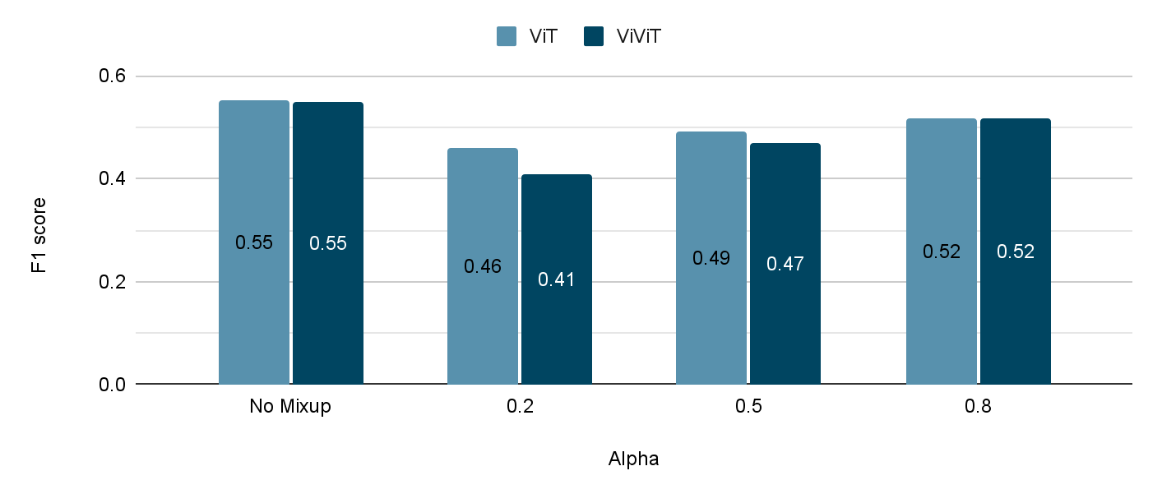}

\caption{Performance (F1 score) of ViT and ViViT models fine-tuned with Mixup augmentation.}
\label{fig:mixup}
\end{figure}

The fourth experimental step augments 20\% of the dataset with the Mixup technique, with three different $\alpha$ configurations. Mixup, even with the additional restriction of combining images belonging to the same patient, fails to contribute to the model's performance even with its best parameter ($\alpha$ = 0.8) for the ViT model (F1 score 0.52) and ViViT model (F1 score 0.52), as shown in Figure~\ref{fig:mixup}.

\newpage
\subsection{Comparisons with previous works}

To our knowledge, recent work by Rudovic et al.~\cite{rudovic2021personalized} is the state-of-the-art for automated binary pain detection on the F1 score metric. Their experimental setup uses a CNN baseline (CDL) but focuses on federalized learning (PFDL), achieving its best-performing model with this technique. As can be seen in Table~\ref{tab:results} our method not only achieves better performance on the F1 score metric with our best performing ViT and ViViT models, but their method is also outperformed by our ViViT-2 model, which trades off performance for more consistent results across folds.

\begin{table}[h]
  \caption{Model results on the UNBC-McMaster dataset}
  \label{tab:results}
  \centering
  \begin{tabular}{lcc}
  \toprule
    Model Name     & F1 score       & AUC \\
    \midrule
    CDL~\cite{rudovic2021personalized}    & 0.46  $\pm$  0.18 &    -  \\
    PFDL~\cite{rudovic2021personalized}   & 0.47 $\pm$   0.20   & - \\
    SPTS + CAPP~\cite{lucey2011painful} & - & 0.84 \\
    SPTS + SAPP + CAPP~\cite{lucey2010automatically} \ & - & 0.85 \\
    \midrule
    ViT-1  & \textbf{0.55} $\pm$ 0.15 &  \textbf{0.88} \\
    ViViT-1    & \textbf{0.55} $\pm$   0.13  & 0.86 \\
    ViViT-2     & 0.49  $\pm$ \textbf{0.04} &  0.76 \\

    \bottomrule
  \end{tabular}
\end{table}

The F1 score is affected by the skew in the labels but AUC is not~\cite{jeni2013facing}. Given that our labels are highly imbalanced, we also report AUC values and compare our results with the works that also report AUC. We compare our top-performing models ViT-1 (AUC 0.88) and ViViT-1 (AUC 0.86) against SPTS + CAPP (AUC 0.84)~\cite{lucey2011painful} and SPTS + SAPP + CAPP (AUC 0.85)~\cite{lucey2010automatically}, and find them to outperform previous works despite not being optimized for this metric.

\subsection{Qualitative Analysis}

\begin{figure}[h] \centering
\includegraphics[width=0.2\linewidth]{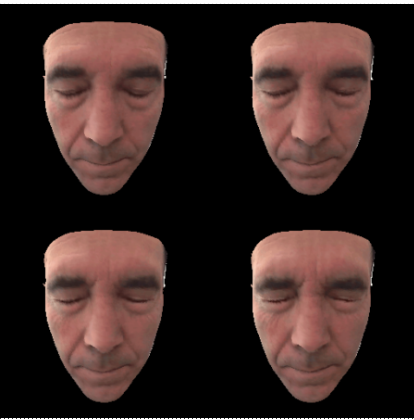}

\caption{The pain sample used to qualitatively analyze the performance of ViViT-1. The first frame in the top-left has a PSPI score of 0, while the remaining three have a PSPI score of 10.}
\label{attentionoriginal}
\end{figure}

Previous works have shown that attention maps can be used to generate visual interpretations for the predictions of vision transformers~\cite{han2021transformer,wang2021progressive}. To demonstrate this feature we will perform a qualitative analysis of the attention maps of ViViT-1 given the sample visible in Figure~\ref{attentionoriginal}, whose pain label is correctly predicted by the model.

\begin{figure}[h] \centering
\includegraphics[width=0.75\linewidth]{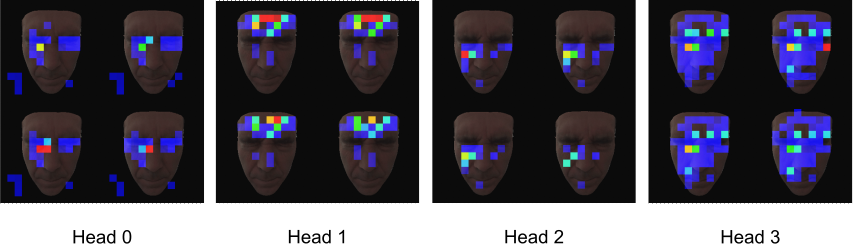}
\caption{Attention maps of individual heads of the final attention layer. Disentangled representations similar to AUs are present in heads 0-2, while head 3 captures most of the face. The activations are thresholded between 0.7 (blue) and 1 (red).}
\label{attentionLL}
\end{figure}

As shown in Figure~\ref{attentionLL}, the last attention layer of ViViT-1 has disentangled representations across its attention heads. These representations partially overlap with AU43 (eyes closed, head 0), AU4 (brow-lowering, head 1)  AU6-7 (orbital tightening, head 2), while others capture a large area of the face (head 3).  

\begin{figure}[h] \centering
\includegraphics[width=0.37\linewidth]{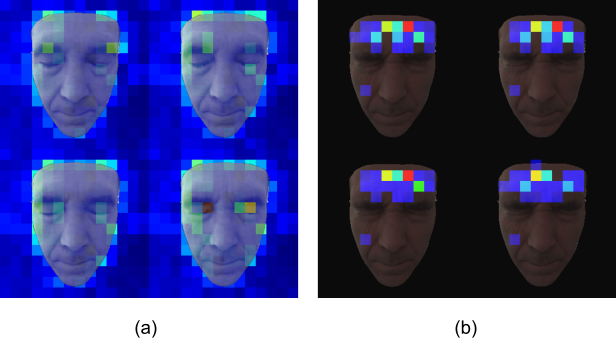}
\caption{Attention maps obtained with max rollout (a) and maximum values of the last layer (b). Frame (b) is thresholded between 0.7 (blue) and 1 (red).}
\label{attentionMM}
\end{figure}

In Figure~\ref{attentionMM}a, the maximum value of the attention patches for the ViViT-1 model is shown, obtained with attention rollout~\cite{attentionrollout}. Attention rollout is a transformer technique that combines information from every attention layer, capturing its flow through the model. In Figure~\ref{attentionMM}b, we show instead the combined maximum values of the heads of the last attention layer for ViViT-1. While the strongest activations are found in the forehead and cheek area for the final layer (b), the flow of information instead clearly originates from the inner brow, lip corner and cheek area (a), which are areas of significance according to facial pain assessment literature~\cite{ekman1978facial}.

The model is clearly capable of generating intrinsic and plausible interpretations for its predictions. Facial regions having higher attention weights in the attention maps are the ones that show a change in appearance and shape during several actions that are observed during a painful expression. It shows that the model effectively detects pain from actions of relevant facial regions.

\section{Discussion}
\label{discussion}

\paragraph{Number of unfrozen layers} The pre-trained transformer model is overall successful across a large variety of parameters for the task of pain detection, contrary to earlier findings on the topic~\cite{xu2021deep}. While the models perform consistently no matter the number of layers, the region around 6 and 12 layers stands out as the better choice both for ViT and ViViT, warranting a deeper investigation of similar parameters and setting a precedent for future work. Despite sharing the top two configurations, the vision and video vision transformers perform best with different numbers of layers, distinguishing their configurations for the following steps.

\paragraph{Learning rate} Comparably, learning rate proves to be a far more delicate parameter, with two of the configurations achieving the worst performance overall across all experiments for ViT and ViViT. Learning rates lower than 2E-03 are generally high-performing, with ViT peaking around 2E-05, achieving the best model performance across all configurations, and ViViT performing best with 2E-04, a much larger learning rate. 

Furthermore, ViViT scores its second best performance at the learning rate of 2E-06 with an extremely low standard deviation across folds, a sign of good generalization. Consistency is a desirable trait for all models, and even more so for delicate tasks in the medical field. Therefore, while achieving only the second best performance according to the metric chosen for this experiment, it possesses a desirable trait and indicates a second region of interest for the tuning of this parameter.

\paragraph{Sharpness-Aware Minimization and Mixup} The SAM optimizer and Mixup augmentation fail to improve the model's performance across a variety of configurations for both ViT and ViViT. While SAM decreases the standard deviation across folds for the ViViT model, it does so by affecting all folds negatively rather than pushing their performance towards an average. Mixup appears ineffective in generating meaningful samples for the model to learn from despite the constraints applied, perhaps due to the delicate nature of FACS and PSPI encoding.   

\paragraph{ViT and ViViT} ViViT fails to outperform ViT despite the strong case for facial dynamics in pain detection literature. While the configurations we use may be limiting the performance of the model, we believe that ViViT models which compute spatio-temporal attention separately,  such as the other ViViT implementations described in the original paper~\cite{arnab2021vivit} might suit this task better. Separating the temporal and spatial attention would benefit the model by allowing larger sequence lengths while avoiding quadratically increased computational time. Furthermore, spatially or temporally-local computed attention might track better the delicate facial dynamics necessary for pain detection.

\paragraph{Societal Impact} To our knowledge, the negative impact of this research is limited to the physical resources used to train and run the models. The presented models should not be used in a clinical context, for which they are untested.

\paragraph{Hardware and Training Time} Training has been performed with a variety of GPUs on Google Colab and Kaggle. Total training time for all 14 configurations is around 30 hours for ViT on Kaggle and around 80 for ViViT on Google Colab.

\section{Conclusion}
\label{conclusion}

In this paper, we have used vision and video vision transformers trained on the UNBC-McMaster dataset for binary pain detection. We have achieved state-of-the-art performance using the F1 score, identified regions of interest in the transformer hyperparameter space, compared the performance of vision and video transformers on this task, and obtained intrinsic plausible interpretations for the performance of the model. Results show that pre-trained transformers can be applied toward pain assessment with good results, after a single epoch of training and on a small unbalanced dataset. Future work could include different augmentation techniques, leave-one-patient-out validation, longer sub-sequences, and more efficient transformer architectures.

\medskip

\bibliographystyle{ieeetran}
\bibliography{bibliography}

\appendix

\end{document}